%% file: template.tex
\documentclass{Interspeech2024}

\usepackage{multirow}
\usepackage{xurl}

\interspeechcameraready 

\title{Just Because We Camp, Doesn't Mean We Should: The Ethics of Modelling Queer Voices.}

\name[affiliation={1*}]{Atli}{Sigurgeirsson}
\name[affiliation={2*}]{Eddie L.}{Ungless}

\address{
  $^1$The Centre for Speech Technology Research, University of Edinburgh, Scotland\\
  $^2$School of Informatics, University of Edinburgh, Scotland}
\email{s2063518@ed.ac.uk, e.l.ungless@ed.ac.uk}


\begin{document}

\maketitle

\begin{abstract}
    Modern voice cloning models claim to be able to capture a diverse range of voices. We test the ability of a typical pipeline to capture the style known colloquially as ``gay voice'' and notice a homogenisation effect: synthesised speech is rated as sounding significantly ``less gay'' (by LGBTQ+ participants) than its corresponding ground-truth for speakers with ``gay voice'', but ratings actually increase for control speakers. Loss of ``gay voice'' has implications for accessibility. We also find that for speakers with ``gay voice'', loss of ``gay voice'' corresponds to lower similarity ratings.
    
    However, we caution that improving the ability of such models to synthesise ``gay voice'' comes with a great number of risks. We use this pipeline as a starting point for a discussion on the ethics of modelling queer voices more broadly. Collecting ``clean'' queer data has safety and fairness ramifications, and the resulting technology may cause harms from mockery to death. 
\end{abstract}

\def\thefootnote{*}\footnotetext{These authors contributed equally to this work}\def\thefootnote{\arabic{footnote}}

\section{Introduction} \label{sec:intro}
\input{sections/01_intro}
\section{Background} \label{sec:background}
\input{sections/02_background}

\section{Case Study}\label{sec:play}
\input{sections/03_toy}

\section{Ethics of Real-world Applications}\label{sec:ethics}
\input{sections/04_ethics}

\vspace{-3pt}
\section{Conclusion} \label{sec:conclusion}
\input{sections/05_conclusion}

\newpage

\section{Acknowledgements}
Both authors are supported
by the UKRI Centre for Doctoral Training in Natural Language Processing, funded by the UKRI (grant EP/S022481/1) and the University of Edinburgh, School of Informatics. Atli Sigurgeirsson is additionally support by Huawei.

\def\url#1{}
\bibliographystyle{IEEEtran}
\bibliography{mybib,eddiezotero}

\end{document}

%% file: sections/01_intro.tex
The controversial Stanford paper \cite{wang_deep_2018} predicting sexuality from dating profile images was intended to raise the alarm on the dangers of deep learning techniques being used to classify people by sexuality \cite{murphy_why_2017}. The significant (and justified) push-back this paper received demonstrates the importance of exploring such issues with great sensitivity. We ask, what does it mean to model queer data ethically? Specifically, we use Text-To-Speech (TTS) synthesis for the speaking style known colloquially as \textit{gay voice} as a case study to initiate a discussion on the ethical implications of modelling queer voice data. 

We use the phrase \textit{gay voice} to refer to the phenomenon of ``phonetic practices'' \cite{mack_influence_2012} typically associated by listeners with gay men\footnote{Bisexual men can also have \textit{gay voice}.}. (Negative) stereotypes of how gay men talk abound, but there is evidence that pitch range, sibilant duration \cite{levon_sexuality_2007} and quality \cite{mack_influence_2012} all influence \textit{perceived speaker sexuality}, although there is no one-to-one correspondence between features and perceived sexuality \cite{mack_influence_2012}.

We focus our case study on voice cloning because this technology has clear ethically significant benefits: those who cannot reliably communicate by voice can use it to maintain their vocal identity and access services. Many companies, for example Microsoft \cite{wang2023neural} and Apple\footnote{\url{https://machinelearning.apple.com/research/personal-voice}}, claim to be able to accurately clone diverse voices using limited source speech; we seek to establish if this likely holds for men with \textit{gay voice}. Failing to accurately model speakers' voices because they are gay is clearly discrimination. However, we discuss why developing a model which captures \textit{gay voice} is also problematic, as part of a larger discussion on the ethics of modelling queer voices.  

We present a brief literature review on \textit{gay voice} and perceived speaker sexuality, necessary background to our hypotheses. We then present a case study of a TTS pipeline which we evaluate using LGBTQ+ participants. We use our experience of designing this pipeline as a ``jumping-off point'' for our ethical discussion. Whilst we find that the model is poor at capturing \textit{gay voice}, we caution that improving our ability to synthesise queer voices comes with many risks. Our contribution is to forefront nuanced discussions of ethics in a discipline that typically prioritises progress over other values \cite{birhane_values_2022}: just because we can do something, doesn't mean we should.

%% file: sections/02_background.tex
\subsection{Gay Voice and Perceived Speaker Sexuality}
Many people hold stereotypes about what gay men sound like: ``soft'', with a ``lisp'', feminine (see \cite{mack_influence_2012}). Without affirming these judgements, we acknowledge that there is evidence of phonetic features that differ between gay and non-gay speakers, some of which influence perceived speaker sexuality. No single (set of) feature(s) is always present in \textit{gay voice} \cite{mack_influence_2012, smyth_male_2003}, but rather gay male speakers are more likely to exhibit some particular ``phonetic practices'' \cite{mack_influence_2012} e.g. hyper- \cite{munson_influence_2006} and mis-articulated /s/ \cite{mack_influence_2012}; longer sibilant duration and greater pitch range also impacts perceived gay sexuality \cite{levon_sexuality_2007}. 

Discourse type and topic may also impact perceived sexuality. Straight men were more likely to be rated as sounding gay when reading scientific texts compared to dramatic texts or with spontaneous speech: however, the same study found no impact of discourse type on perceptions of speaker sexuality for gay speakers \cite{smyth_male_2003}. A recent study \cite{kachel_queer_2023} found the difference in perceived sexuality between straight and non-straight speakers was greatest when speakers were discussing a topic unrelated to sexuality (in a language unknown to listeners). This was because (perhaps counter-intuitively) gay speakers were rated as less gay when discussing LGBTQ+ topics. We speculate that when attention is focused on their identity, internalised homophobia and associated stigma \cite{smyth_male_2003} may lead gay speakers to subconsciously suppress \textit{gay voice}. 

\vspace{-5pt}
\subsection{Hypothesis Development}
We ask: can perceived \textit{gay voice} be successfully modelled by end-to-end voice-cloning TTS models (Research Q1)?
Any speaker identity modelling necessarily involves some loss, and we believe this will be exacerbated in the case of men with \textit{gay voice.} Existing speech datasets are not focused on capturing queer voices, thus men with \textit{gay voice} will be vastly outnumbered, if represented at all. Therefore we believe that because these models implicitly do not model \textit{gay voice}, the resulting output will be poor in terms of fidelity to source voice. 

As such, we predict:
(H1) Compared to the difference between ground truth and synthesised control speaker utterances, synthesised \textit{gay voice} utterances will show a larger decrease in perceived gayness compared to ground truth utterances. 

Because we believe perceived gayness is an important part of speaker identity, we also believe that this loss in perceived gayness will be associated with a loss in perceived speaker similarity. As such, we predict: (H2) Compared to the difference between ground truth and synthesised control speaker utterances, synthesised \textit{gay voice} utterances will be rated as having a lower speaker similarity with ground truth utterances. 

Finally, we will explore which steps in a speech synthesis pipeline will have the biggest impact on the fidelity of ``gay voice'' (Research Q2).

\vspace{0.2em}
\noindent \textbf{\textsc{reflexivity statement}}: We approach this work as authors who identify as gay men. One author's experience as a trans gay man has given him lived experience of deliberately trying to achieve \textit{gay voice}. We are white and most familiar with \textit{gay voice} in White Aligned English, which may have influenced our initial choice of speakers.

%% file: sections/03_toy.tex
\subsection{Method}\label{sec:method}

\textbf{\textsc{stimuli creation}:}
There are no available TTS corpora labelled by sexuality that we are aware of. We resort to selecting speakers with \textit{gay voice} from a large generic speech corpus - Ted-Lium 3 \cite{hernandez2018ted} - which includes speech from several self-identifying gay men. All were speakers of US American English - this narrow scope ensured familiarity for ourselves as data curators, and our annotators. We manually selected speakers based on our shared perception of them having \textit{gay voice} (henceforth, \texttt{gay} group). Speakers were included if both authors agreed the speaker had \textit{gay voice}, although the perceived ``strength'' of \textit{gay voice} varies across selected speakers. We confirmed that all selected speakers publicly identify as gay and/or are in public relationships with men. 

For our control data, we chose speakers who we perceived not to have \textit{gay voice} (henceforth, \texttt{control} group) and who do not publicly identify as gay, acknowledging that these speakers may nonetheless be gay: their ``real'' sexuality is not directly relevant to whether \textit{gay voice} is captured by speech synthesis models. We chose approximate controls for each \textit{gay voice }speaker: similar age and recording year; we paired Black \texttt{control} speakers with Black \texttt{gay} speakers. We did not control for regional accent variation, talk venue (which impacts quality) or talk topic, which may influence \textit{gay voice} perception \cite{kachel_queer_2023}. Our data consists of speakers  aged 29-58 at time of recording. We include 20 white and 4 Black speakers (perceived race). To the best of our knowledge, all speakers are men.

We study the perception of \textit{gay voice} at different stages of TTS development. We consider two different raw conditions (henceforth, clip types) for evaluation. First, long utterances: we select three 30 second segments (henceforth \textbf{Segment}) per speaker, ensuring none contained discussion of LGBTQ+ topics. TTS models are typically not trained on such long utterances, but we anticipate that longer segments will better capture speaker vocal identity. We then consider ground-truth segmented utterances (henceforth \textbf{Raw-utt}), as segmented in TED-Lium 3, which reflect typical training data. We select 15 such utterances per speaker, excluding any which contain terms from a list of queer vocabulary (comprised of a list taken from Wikipedia\footnote{\url{https://en.wikipedia.org/wiki/LGBT_slang}} and supplemented through our own knowledge) to minimise the likelihood of sexuality disclosure. 

The Ted-Lium 3 corpus is not purposefully made for TTS modelling, and many utterances include background noise, music and unlabelled speech. For our third clip type, we employ the SepFormer voice-enhancement model \cite{subakan2021attention} trained on the Microsoft-DNS 4 voice-enhancement challenge corpus \cite{dubey2022icassp}. This pretrained model is accessed through the \textit{SpeechBrain} toolkit \cite{speechbrain}. We pass all  \textbf{Raw-utt} through the voice enhancement model to form our third clip type, \textbf{VE-utt}. 

We finally consider the perceived \textit{gay voice} of synthesised speech. The data we collected was insufficient to train a TTS model from scratch. Instead, we use a pretrained version of XTTS-v2 from Coqui \cite{coqui}. XTTS-v2 is a popular multi-lingual, multi-speaker TTS model that supports voice cloning. The model is trained on more than 16,000 hours of publicly available speech data and supports zero-shot reference-based voice cloning. That is, given a reference utterance, the model can estimate a speaker embedding which is then used to condition speech generation. We evaluate two different clip types of synthesised speech: 1) \textbf{Copy-synth}, where the target text matches the text in the reference (\textbf{VE-utt}) used to generate the speaker embedding, and 2) \textbf{Synth} where the target text does not match. We anticipate that \textbf{Copy-synth} represents the optimal conditions for models such as XTTS-v2 to model a speaker's identity.

\vspace{0.5em}
\noindent \textbf{\textsc{participants:}}
All our participants were US and UK L1 speakers of English. We only recruited LGBTQ+ participants because we are principally interested in whether synthesised \textit{gay voice} is perceptible to other community members, not whether it is detectable by out-group members, and listener sexuality may influence perceived sexuality \cite{smyth_male_2003,munson_lavender_2011}.
\vspace{0.5em}
\noindent \textbf{\textsc{metrics:}}
Metric design for perceived \textit{gay voice} differs across the literature in ways that are likely to be significant \cite{munson_lavender_2011}. We chose our specifications as follows: as we are interested in perception of \textit{gay voice} not perceived speaker sexuality, we ask participants whether the voice \emph{sounds} gay rather than whether the speaker \emph{is} gay, similar to \cite{mack_influence_2012} but counter to e.g. \cite{kachel_queer_2023}. We employ a 7-point Likert scale (in line with e.g. \cite{gaudio_sounding_1994, levon_sexuality_2007, kachel_queer_2023}) where 1 is ``definitely sounds straight'' and 7 is ``definitely sounds gay''. We asked participants to rate naturalness on a 5-point Likert scale (1 ``bad'', 5 ``excellent''). Similarity is rated on a scale of 0-100 (0 ``completely different'', 100 ``exactly the same''), between a reference \textbf{Segment} and \textbf{Synth}. 

\vspace{-4pt}
\subsection{Pre-study}
We conducted a pre-study to confirm that our \texttt{gay} and \texttt{control} speakers showed divergent \textit{gay voice} ratings. We recruited 13 participants to rate 30 \textbf{Segments} each, such that each \textbf{Segment} was rated approximately 5 times. Based on participant ratings, we discarded 5 speakers from each group, leaving those with a rating above 5 (for \texttt{gay}) or below 3 (for \texttt{control}). 

\vspace{-4pt}
\subsection{Study One}
\textbf{\textsc{procedure}:} To test whether \textit{gay voice} is lost in speech synthesis (RQ1: H1) and at which stage in the pipeline \textit{gay voice} is lost (RQ2), each participant rated (a) two 30-second \textbf{Segments} and (b) a mixture of 30 utterances (\textbf{Raw-utt}, \textbf{VE-utt} and \textbf{Copy-synth}) for how gay the voice sounded. This was repeated for (c) \textbf{Segments} and (d) utterances (\textbf{Raw-utt}, \textbf{VE-utt} and \textbf{Copy-synth}) for how natural the clips sounded. All clips were allocated at random, and the order of tasks (a-d) was randomized. Two participants were discarded for failing attention checks. Each utterance was rated approximately 4 times.

\begin{figure}[!h]
    \centering
\includegraphics[width=\linewidth]{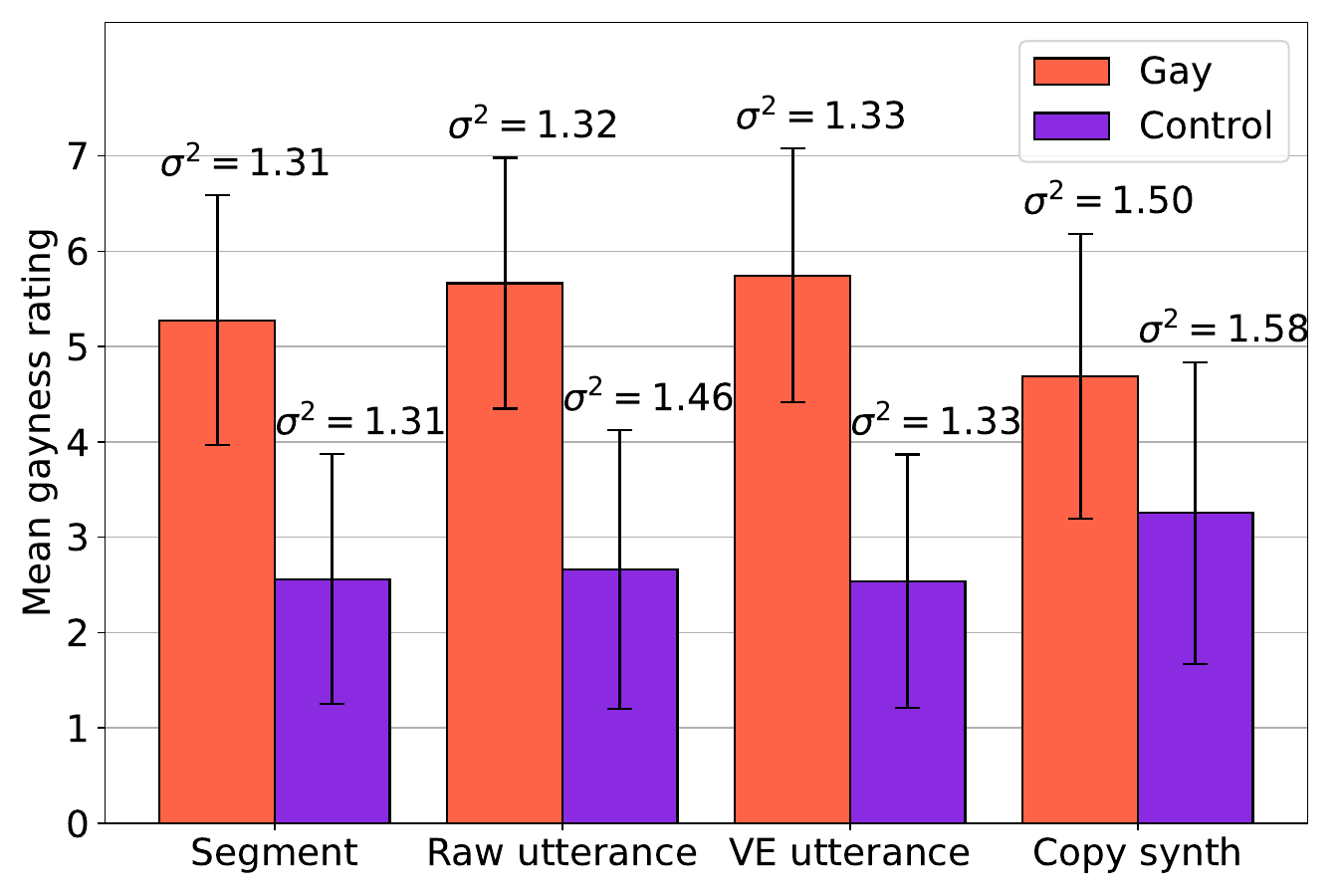}
    \vspace{-20pt}
    \caption{Mean \textnormal{gay voice }rating of both speaker groups in each clip type.}
    \label{fig:mean-gayness}
    \vspace{-15pt}
\end{figure}

\vspace{1.2em}
\noindent \textbf{\textsc{results and discussion}:} Results for Study One are given in Table \ref{tab:results} and visualised in Figure \ref{fig:mean-gayness}. In R we used \texttt{lme4} to produce linear mixed effect models, and \texttt{afex} to conduct likelihood ratio testing for $p$-values (appropriate given nested random variables with many levels). As is to be expected, \texttt{control} speakers were rated as having much lower \textit{gay voice} than \texttt{gay} speakers: the estimated effect was $-2.59 \pm 0.26$ per a linear mixed effects model with group as fixed effect, and annotator, speaker and utterance ID as random effects. This was significant, $\chi^2(1) = 26.02, p<.001$. 

Comparing mean ratings for \textbf{Segments} versus \textbf{Copy-synth} for \texttt{gay} speakers we find that \textbf{Copy-synth} were rated as sounding much less gay. Unexpectedly, ratings \emph{increased} for \texttt{control} speakers. We speculate that this may be because for straight men, read speech of the kind typically used to train TTS technologies, is rated as more gay sounding than spontaneous speech or when reading ``dramatic'' texts \cite{smyth_male_2003}. 

\input{tables/main_results}

\noindent A linear mixed effects model of \textit{gay voice} ratings for \textbf{Segments} and \textbf{Copy-synth}, with group and clip type as fixed effects, and annotator, speaker and utterance ID as random effects, found a main effect of group, estimated as $-2.68 \pm 0.38$ ($\chi^2(1) = 18.52, p<.001$), a non-significant effect of clip type and a significant interaction ($1.20 \pm 0.31$, $\chi^2(1) = 14.26, p<.001$). In a sense, our prediction that \texttt{gay} speakers would show a larger decrease in ratings \textbf{(H1) is supported}, in that ratings actually increased for \texttt{control} speakers. 

Looking at average \textit{gay voice} rating by clip type to explore RQ2, we observe that ratings actually increase between \textbf{Segment} and \textbf{Raw-utt}, particularly for \texttt{gay} speakers. This is likely due to sexuality disclosure despite attempts to prevent this. There does not seem to be a substantial difference between \textbf{Raw-utt} and \textbf{VE-utt}, for either group. There is a large drop in ratings between \textbf{VE-utt} and \textbf{Copy-synth} for \texttt{gay} speakers and an increase for control speakers. 

We conducted a series of pairwise linear mixed effect models with clip type and group as fixed effects and annotator, speaker and utterance ID as random effects. We found that between \textbf{Segment} and \textbf{Raw-utt}, clip type has a weak significant effect estimated as $0.44 \pm 0.20$, $\chi^2(1) = 3.91, p=.048$ (main effect of group is $-2.71 \pm 0.38 $, $\chi^2(1) = 23.65, p<.001$;  non-significant interaction). Between \textbf{Raw-utt} and \textbf{VE-utt} there is a non-significant impact of clip type (main effect of group is $-3.03 \pm 0.25$, $\chi^2(1) = 28.38, p<.001$; non-significant interaction). Between \textbf{VE-utt} and \textbf{Copy-synth}, there is a significant effect of clip type, estimated as $-1.49 \pm 0.23$ ($\chi^2(1) = 4.50, p=.034$); a main effect of group ($-1.49 \pm 0.23$, $\chi^2(1) = 25.38, p<.001$); and a significant interaction effect ($1.75 \pm 0.16$, $\chi^2(1) = 115.38, p<.001$). In answer to \textbf{RQ2, synthesis has the biggest impact on \textit{gay voice} ratings}, over using short, out-of-context clips, and voice enhancement.

As is to be expected, perceived naturalness decreases through the pipeline, the biggest drop coming at synthesis. \textit{Gay voice} speakers were rated as less natural on average, which we speculate is due to greater background noise.

\vspace{-8pt}

\subsection{Study Two}
\textbf{\textsc{procedure}:} To test whether loss in perceived gayness is associated with a loss in similarity (RQ1: H2) we presented participants with a 30-second \textbf{Segment} for a speaker as reference, plus a corresponding \textbf{Synth} for that speaker alongside two anchor utterances from other speakers (one each from \texttt{gay} and \texttt{control}), and asked them to rate the utterances on similarity. Similarity accuracy was measured as how often the correct sample was selected as the most similar. \textbf{Segments} and \textbf{Synth} were separately rated on naturalness and \textit{gay voice}. Each participant rated 20 speakers' utterances; each utterance was rated 4 times. Clips were randomly allocated. Task order was randomised. 1 participant was discarded for failing attention checks.

\vspace{0.5em}
\noindent \textbf{\textsc{results and discussion}:} We conduct a Kruskal-Wallis test (as our residuals are not normally distributed) and find that there is no significant difference in similarity accuracy between the groups, so \textbf{RQ1:H2 is not supported}. However, when we examine the correlation between \textit{gay voice} rating and similarity, we notice for \texttt{gay} speakers there is a medium correlation between size of loss of \textit{gay voice} and similarity accuracy, $r_s=-3.44, p=.021$, and no such correlation for \texttt{control} speakers. It seems loss of perceived \textit{gay voice} for \texttt{gay} speakers is more pertinent to similarity judgements than ``loss'' of ``straight voice''.

\begin{figure}[!h]
    \centering
\includegraphics[width=\linewidth]{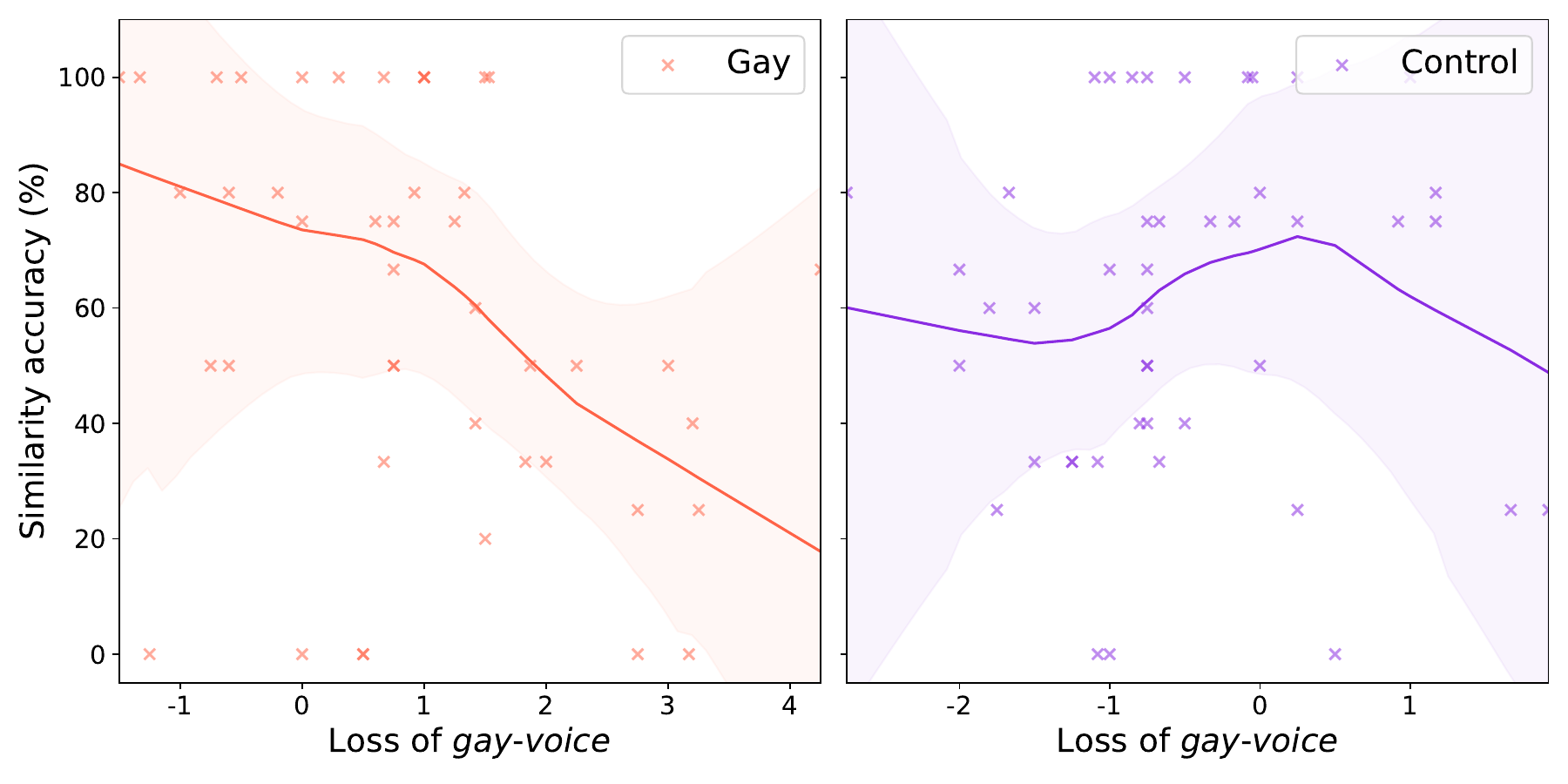}
    \vspace{-17pt}
    \caption{Graphs showing the correlation between the similarity accuracy and reduction in gay voice between reference \textbf{Segments} and \textbf{Synth} (A positive value indicates a loss). Lines represent LOWESS-smoothing and shaded area indicates the 95\% confidence intervals. }
    \label{fig:sim-vs-accuracy}
    \vspace{-15pt}
\end{figure}

%% file: tables/main_results.tex
\begin{table}[h]
\centering
\caption{Mean rating $\pm$ standard deviation for naturalness and for \textnormal{gay voice}, by condition and by group.} \label{tab:results}
\resizebox{\linewidth}{!}{%
\begin{tabular}{lcccc}
\toprule
                       & \multicolumn{2}{c}{\textbf{Naturalness}} & \multicolumn{2}{c}{\textbf{\textit{Gay Voice}}} \\
                       & \texttt{Gay}      & \texttt{Control}      & \texttt{Gay}    & \texttt{Control}   \\
\midrule
\textbf{Seg.}       & $3.6\pm1.2$         & $4.2\pm0.8$         & $5.3\pm1.3$       & $2.6\pm1.3$      \\
\textbf{Raw-utt}   & $3.5\pm1.2$         & $3.9\pm1.0$         & $5.7\pm1.3$       & $2.7\pm1.5$      \\
\textbf{VE-utt}    & $3.4\pm1.2$         & $4.0\pm1.0$         & $5.7\pm1.3$       & $2.5\pm1.3$      \\
\textbf{Copy-S.}         & $2.7\pm1.3$         & $2.9\pm1.3$         & $4.7\pm1.5$       & $3.3\pm1.6$\\
\bottomrule
\end{tabular}%
}
\end{table}

%% file: sections/04_ethics.tex
Having established that our case study pipeline is not able to fully capture \textit{gay voice}, we now discuss the ethical implications of building a speech synthesis model that can model queer voices. We consider the so called ``dual use'' of this technology  \cite{mohammad_ethics_2022}. Specifically, we briefly overview the positive uses of this technology to demonstrate why researchers might be interested in pursuing this goal, before considering key ethical issues this pursuit raises. Throughout, we extend beyond speech synthesis to other tasks. Modelling queer voices should be approached with extreme caution due to the potential for harm.

\subsection{Potential Positives}
In addition to differences between gay and straight male speakers discussed in Section \ref{sec:background}, queer speakers may differ from non-queer speakers for a number of social and physiological reasons. For example, lower F0 may reflect transmasculine identity \cite{leann_bright_2022}, and transmasculine people undergoing testosterone therapy may experience the vocal quality of ``entrapment'' (though cf. \cite{cler_longitudinal_2020}) and wish to maintain this in voice cloning. Given use of speech synthesis technologies to model a variety of accents, it seems natural to extend these efforts to capturing queer voices. This seems particularly vital in cases where voice cloning or speech synthesis is adopted by people who are disabled, who may otherwise have an important aspect of their identity erased by technology that fails to capture queer voices. Beyond voice cloning, voice gender recognition (VGR) systems not trained on diverse data may misgender speakers, which can lead to problems e.g. in speech translation, although we question if VGR is ethically justified in the first place \cite{hamidi_gender_2018}.

\subsection{Data Subjects and Data Security} 
Having a dataset of private individuals labelled for sexuality could be harmful if the data is made public, as individuals (recognisable by their voice) may be ``outed'' - in many countries this has potentially fatal consequences (similar concerns have been expressed for image data \cite{ungless_stereotypes_2023}). However, using only public figures would severely limit the amount and diversity of data available. That no database of voices labelled by sexuality currently exists may reflect such safety concerns (although this may also reflect a lack of interest in modelling queer voices by a research community that is primarily not queer). \cite{markl_mind_2022} cautions us to consider ``what is missing from a dataset'' when considering power inequalities. Beyond gay men's voices, it is likely that existing databases are missing diverse queer voices. Traditional recruitment methods may fail to reach these communities, and marginalised groups are often not given prestigious platforms that are systematically recorded like TED. Acquiring diverse data can be problematic, given the relative power of developers versus data subjects \cite{ungless_stereotypes_2023,fussell_how_2019, mahelona_openais_2023}.
\subsection{Cleaning Queer Data}
The process of cleaning queer data for modelling can raise ethical issues. Whilst the voice enhancement tool we used did not impact perceived \textit{gay voice}, others we discounted\footnote{in order to produce the most accurate synthesised voices} seemed to have a more substantial impact. Preservation of queer voices is likely not accounted for when these tools are developed. Features of queer speech like entrapment may be processed as noise to be cleaned - analogous to the treatment of queer people as undesirable deviations from the norm.

Although we do not include VGR in our pipeline, it is likely to be inaccurate for gender diverse people, as noted above. If annotation is required, annotators from outwith the queer community may provide less accurate labels as they are removed from the associated culture. Whilst it is not clear whether gay listeners are significantly better at recognising gay voice than straight listeners \cite{smyth_male_2003,munson_lavender_2011}, and further listeners are able to recognise \textit{gay voice} in speakers of other languages \cite{kachel_queer_2023}, the fact that so much annotation of Western data happens in the Global Majority may introduce cross-cultural artefacts into the accuracy of labels \cite{smart_discipline_2024}. 

\subsection{Misappropriation of Queer Voices}
Speech synthesis technologies, including TTS and speech-to-speech, could be used to misappropriate and mock queer voices. For example, \textit{gay voice} may be used to narrate doing something which is stereotypically associated with gay people, such as being bad at sports, weak \cite{rolle_homonegativity_2022}, promiscuous or predatory \cite{simon_relationship_1998}. TikTok, the popular social media platform, seems a plausible arena where this might occur: by virtue of its design, which encourages the meme-ification of audio clips, TikTok is already implicated in the mis-appropriation of different marginalised identities through voice \cite{jones_lip-synching_2023,ilbury_recontextualisation_2023}. TikTok is also a platform where the public are encouraged to play with speech synthesis models. It is plausible that models designed to produce synthesised \textit{gay voice}, or other queer voices, may be used to mock queer people. We discourage integration of queer voices into speech synthesis models available to the general public.

\subsection{Risk of Real-life ``Gaydars''} 
A particularly harmful use of models of queer voices would be the creation of ``gaydars'' which can identify queer individuals by voice. Whilst the speaking style we study in this paper is stereotypically associated with gay men, such that a classification model would likely be on par with a person at identifying gay speakers, there are phonetic features listeners do not associate with particularly identities but are nonetheless evident e.g. for bisexual men \cite{morandini_bidar_2023}. Anecdotally, qualities associated with trans men's voices are little known about outwith the trans community. A model may learn these features to create gaydars or ``clockers'' \cite{schiffer_researching_2022}, which could put queer individuals at fatal risk, particularly trans individuals \cite{noauthor_tvt_2015}. We advise against public release of labelled queer voice data, or the results of modelling studies\footnote{for example, we do not release details of the different latent representations for \texttt{gay} and \texttt{control} voices.}, which may inform the development of such dangerous technologies.

%% file: sections/05_conclusion.tex
Though we find that a typical multi-speaker TTS pipeline is unable to accurately capture \textit{gay voice}, we are hesitant to recommend resolving these issues, at least in publicly available models. It can be tempting, when considering a technology in isolation of its social context, to fix any identified bias in the output, typically by sourcing additional data. However, attempts to resolve biases - such as the inability to capture queer voices - cannot come at the expense of the safety of the group we are claiming to help. 